\documentclass[letterpaper, 10 pt, conference]{ieeeconf}  %

\IEEEoverridecommandlockouts                              %

\usepackage{graphics} 
\usepackage{amsmath} 
\usepackage{amssymb}  
\usepackage{hyperref}
\usepackage{svg}
\usepackage{float}
\usepackage{multirow}
\usepackage{array}
\usepackage{xspace}
\usepackage{booktabs}
\usepackage{caption}
\usepackage[flushleft]{threeparttable}
\usepackage{etoolbox}
\usepackage{comment}
\usepackage{adjustbox}
\usepackage[noadjust]{cite}

\usepackage{caption}
\usepackage{subcaption}
\title{\LARGE \bf SELP: Generating Safe and Efficient Task Plans for Robot Agents with Large Language Models}

\author{
    Yi Wu, Zikang Xiong, Yiran Hu, Shreyash S. Iyengar, Nan Jiang, \\ Aniket Bera, Lin Tan, and Suresh Jagannathan  %
    \thanks{ This paper was accepted by ICRA 2025. }
\thanks{ Authors affiliate to Computer Science Department, Purdue University, IN 47906, USA. {\tt\small \{wu1827, xiong84, hu954, iyengar3, jiang719, aniketbera, lintan\}@purdue.edu}, \tt\small \{suresh@cs.purdue.edu\}}
}

\begin{document}
\maketitle

\newcommand{\ourname}{SELP\xspace}

\newcommand{\todoc}[2]{{\textcolor{#1}{\textbf{#2}}}}
\newcommand{\todoblack}[1]{{\todoc{black}{\textbf{[[#1]]}}}}
\newcommand{\todored}[1]{{\todoc{red}{\textbf{[[#1]]}}}}
\newcommand{\todogreen}[1]{\todoc{green}{\textbf{[[#1]]}}}
\newcommand{\todoblue}[1]{\todoc{blue}{\textbf{[[#1]]}}}
\newcommand{\todoorange}[1]{\todoc{orange}{\textbf{[[#1]]}}}
\newcommand{\todobrown}[1]{\todoc{brown}{\textbf{[[#1]]}}}
\newcommand{\todogray}[1]{\todoc{gray}{\textbf{[[#1]]}}}
\newcommand{\todopurple}[1]{\todoc{purple}{\textbf{[[#1]]}}}
\newcommand{\todopink}[1]{\todoc{magenta}{\textbf{[[#1]]}}}
\newcommand{\todocyan}[1]{\todoc{cyan}{\textbf{[[#1]]}}}
\newcommand{\todoviolet}[1]{\todoc{violet}{\textbf{[[#1]]}}}
\newcommand{\todo}[1]{\todored{TODO: #1}}
\newcommand{\todoafter}[1]{\todoorange{TODOAFTER: #1}}

\newcommand{\lin}[1]{\todoblue{Lin: #1}}
\newcommand{\yi}[1]{\todoviolet{Yi: #1}}
\newcommand{\aniket}[1]{\todored{Aniket: #1}}
\newcommand{\yiran}[1]{\todocyan{Yiran: #1}}
\newcommand{\yash}[1]{\todopurple{Shreyash: #1}}
\newcommand{\nan}[1]{\todobrown{Nan: #1}}
\newcommand{\suresh}[1]{\todoorange{Suresh: #1}}
\newcommand{\zikang}[1]{\todopink{Zikang: #1}}

\newcommand{\quotes}[1]{``#1''}
\def\code#1{\texttt{#1}}

\renewcommand{\todoc}[2]{\relax}

\newcommand{\buchi}{Büchi }
\newcommand{\buchiauto}{\mathcal{B}}

\newcommand{\distance}{10pt}
\setlength{\textfloatsep}{\distance} %
\setlength{\floatsep}{\distance} %
\setlength{\intextsep}{\distance} %
\setlength{\dbltextfloatsep}{\distance} %
\setlength{\dblfloatsep}{\distance} %

\begin{abstract}
  Despite significant advancements in large language models (LLMs) that enhance robot agents' understanding and execution of natural language (NL) commands, ensuring the agents adhere to user-specified constraints remains challenging, particularly for complex commands and long-horizon tasks.
  To address this challenge, we present three key insights,
  \emph{equivalence voting}, \emph{constrained decoding}, and \emph{domain-specific fine-tuning}, which significantly enhance  LLM planners' capability in handling complex tasks.
  \emph{Equivalence voting} ensures consistency by generating and sampling multiple Linear Temporal Logic (LTL) formulas from NL commands, grouping equivalent LTL formulas, and selecting the majority group of formulas as the final LTL formula.
  \emph{Constrained decoding} then uses the generated LTL formula to enforce the autoregressive inference of plans, ensuring the generated plans conform to the LTL.
  \emph{Domain-specific fine-tuning} customizes
  LLMs to produce safe and efficient plans within specific task domains. Our approach, \textbf{S}afe \textbf{E}fficient \textbf{L}LM \textbf{P}lanner (\textbf{\ourname}), combines these insights to create LLM planners to generate plans adhering to user commands with high confidence.
  We demonstrate the effectiveness and generalizability of \ourname across different robot agents and tasks, including drone navigation and robot manipulation. For drone navigation tasks, \ourname outperforms state-of-the-art planners
  by 10.8\% in safety rate (i.e., finishing tasks conforming to NL commands) and by 19.8\% in plan efficiency. For robot manipulation tasks, \ourname achieves 20.4\% improvement in safety rate. 
  Our datasets for evaluating NL-to-LTL and robot task planning will be released in \href{https://github.com/lt-asset/selp}{github.com/lt-asset/selp}. 
\end{abstract}

\section{INTRODUCTION}

Recent advancements in large language models have significantly improved robots' abilities to understand and plan given natural language commands~\cite{code_as_policy, saycan, ProgPrompt}. This breakthrough substantially broadens the scope of tasks that robots can autonomously perform with high adaptability across various domains such as autonomous driving~\cite{yang2023survey}, robotics task and motion planning~\cite{jiang2022vima,driess2023palm,saycan}, and human-robot collaboration~\cite{liu2023llm}.
For example, LLM planners can interpret a command such as ``cook a steak and then wash the pan'',  and seamlessly organize this into a plan for cooking followed by cleaning. Combined with prompt-based techniques such as in-context learning and chain-of-thoughts reasoning \cite{saycan, huang2023reasoning,zhu2024knowagent,li2024llms}, LLM planners bring improvements on multiple planning tasks.

\begin{figure*}[h]
    \centering
    \includegraphics[width=0.76\linewidth]{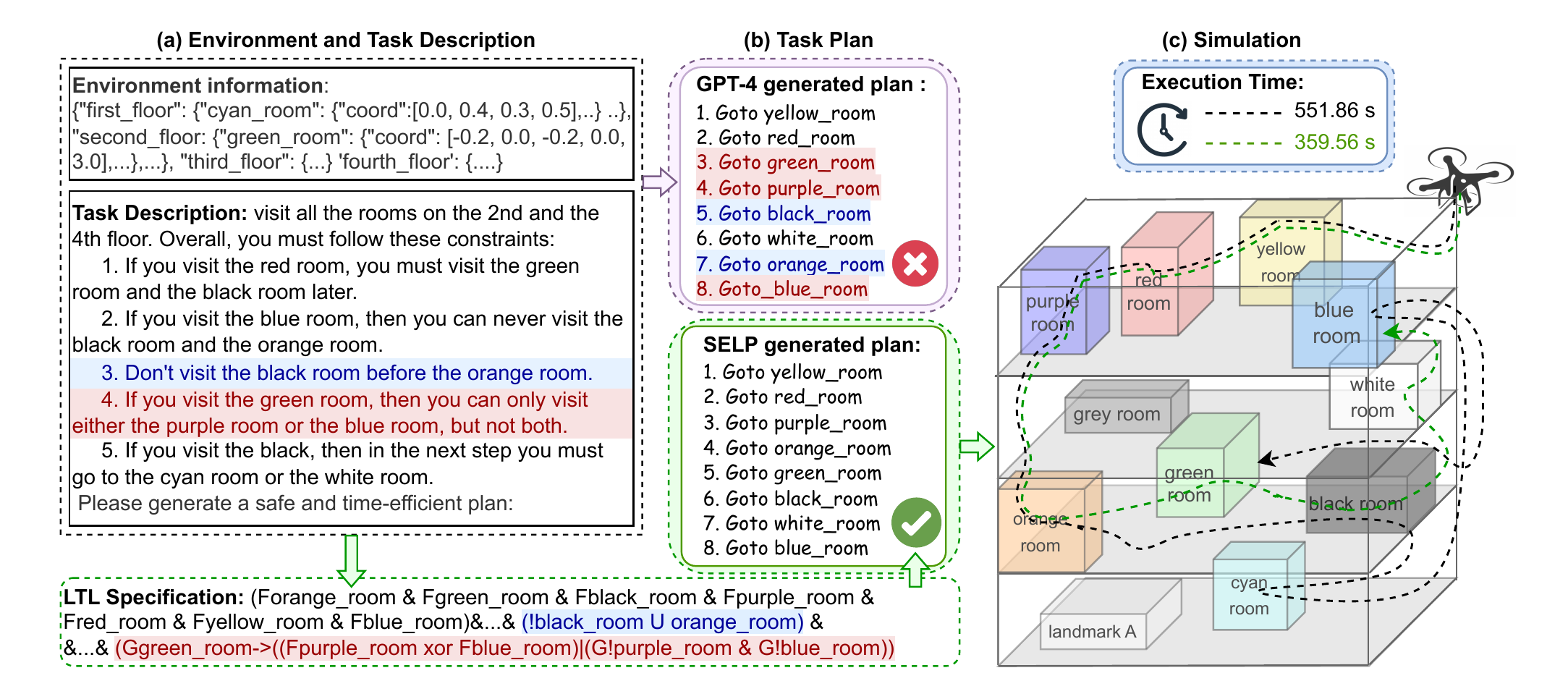}
    \caption{
    Motivating Example: given a drone navigation task (a), \ourname generates a safe and efficient plan (green box in (b)), while GPT-4
        generates \textit{unsafe} plan violating the constraints highlighted in red and blue.
        (c) shows two safe trajectories generated by \ourname from our plan in (b): (1) with domain-specific fine-tuning (green), and (2) without domain-specific fine-tuning (black).  The execution times of the green and black trajectories are 359.56s and 551.86s respectively. }
    \label{fig:Motivation_example}
\end{figure*}

Despite these progresses, LLM planners reach their performance limits as the complexity of commands increases~\cite{huang2023reasoning,zhu2024knowagent,li2024llms}. This complexity manifests in different dimensions: commands may involve intricate logical dependencies with multiple pre- and post-conditions, or tasks may span long time horizons, requiring flawless execution at each step.
Typically, to evaluate a planner’s ability to handle complex tasks, two critical metrics are considered: \emph{safety}, defined as the planner's compliance with given commands, and \emph{efficiency}, measured as the time at which a robot completes a task.
With increasingly more complex commands, existing LLM planners produce more unsafe and inefficient plans, preventing them from being applied to complex or real-world domains. Fig.~\ref{fig:Motivation_example} shows an example where the user requires the drone to visit rooms with some constraints on the visiting order. %
GPT-4 generates an unsafe plan (shown in the purple block) that disobeys the constraints.

Another challenge appears when fine-tuning LLM planners with safety and efficiency objectives. These objectives can sometimes conflict, making it difficult for a model to learn to balance them. Safety often requires conservative planning, incorporating redundancies, and thorough checks to avoid errors, which can lead to longer execution times and lower efficiency. On the other hand, optimizing for efficiency typically involves minimizing the number of steps and the time taken to complete a task, which can increase the risk of unsafe plans.

\ourname effectively addresses these limitations. Similar to the existing technique~\cite{safe-chip}, \ourname starts with translating NL into a set of LTL specifications as an intermediate representation. However, \ourname provides confidence in the correctness of these LTL specifications with a  \emph{equivalence voting} mechanism, which checks the logical equivalence of LTL specifications and selects the majority as the %
specification. The key observation is that an LLM with over 50\% accuracy in generating correct LTL specifications can provide high confidence in correctness through majority voting. %
Then, \ourname directly uses the majority of LTL specification for \emph{constrained decoding} on an LLM planner \cite{saycan, huang2023reasoning,zhu2024knowagent,li2024llms}. The constrained decoding  %
translates LTL specifications into a Büchi automaton that monitors and masks inconsistent tokens, enforcing the LLM to resample until the plan conforms to the given specifications.
Finally, we fine-tune the LLM %
to  boost both efficiency and safety.
For the same example in Fig.~\ref{fig:Motivation_example}, \ourname produces a safe plan (green box), which is also efficient during simulation (the green trajectory in  (c)) with 34.85\% less execution time.

In summary, this paper makes the following contributions:

\begin{itemize}
    \item We propose an equivalence voting mechanism to increase confidence in generating correct LTL specifications from natural language.
    \item We design an LTL-enforced constrained decoding algorithm %
          to prune out unsafe actions. %
    \item We fine-tune LLMs with safe and efficient plans to improve planning safety and efficiency.
    \item We build our approach \ourname that combines the three techniques above, outperforming the best-performing SOTA LLM planner by 11.63\% in safety rate and by 19.78\% in time efficiency.
    \item Additionally, we create two new datasets for evaluating complex drone navigation tasks and tabletop manipulation tasks.
\end{itemize}

\section{RELATED WORK}

\smallskip
\textbf{LLM Agents for Robotics:}
LLMs have shown promising capabilities in robotics scenarios when tasked with agent planning~\cite{ ProgPrompt, huang2022language, vemprala2023chatgpt, huang2022inner, shah2023lm}. Recent work  \cite{saycan,hazra2024saycanpay} applies LLMs for planning with robotic affordances and \cite{code_as_policy, ProgPrompt, tidybot} contribute their novel formulations of using LLMs to generate Python code as robot plans.
A closely related approach to this work is Safety-Chip~\cite{safe-chip}, which proposes a safety constraint module to monitor action sequences generated by LLMs using LTL automata. If an action is unsafe, Safety-Chip re-prompts the LLM to analyze and regenerate an action. In contrast, we employ constrained decoding to effectively prune unsafe actions by directly modifying LLMs' probability distribution. Additionally, we enhance the LLM's planning capability through training rather than relying solely on prompting. ~\cite{autotamp} uses LLMs to generate Signal Temporal Logic (STL) and then employs a solver-based STL planner to generate trajectories, while our work focuses on developing learning-based LLM planner. Other approaches~\cite{xie2023translating, liu2023llm+} convert NL to Planning Domain Definition Language (PDDL) as input for classic planners to generate plans. However, directly using PDDL to solve planning problems limits the ability to improve plan efficiency through fine-tuning and struggles to scale to long-horizon, logic-complex planning problems due to the inherent computational complexity of PDDL solvers.

\smallskip
\textbf{Translating NL to LTL:}
Efforts to convert NL into LTL span from traditional recurrent neural network~\cite{berg2020grounding, nl2ltl_RNN1, oh2019planning} to latest works based on LLMs~\cite{Data-Efficient-Learning, lang2ltl, chen2023nl2tl}. 
However, two common challenges were the contamination of the training or testing datasets with noise and the decreased performance with the increased complexity of NL or LTL. Our strategy resolves these problems by employing LTL grammar and the paraphrasing capabilities of LLMs to create more semantically diverse and complex datasets, enabling more efficient training of LTL translation.

\section{PRELIMINARIES}
\textbf{Linear Temporal Logic:}
Robotic systems frequently employ LTL to formalize complex motion plans and verify task execution \cite{LTLapp1,LTLapp2, DeGiacomo2018FoundationsFR}. The grammar of LTL specification is defined recursively as:
\begin{equation}
\phi := \alpha \mid \neg\phi \mid \phi \land \varphi \mid \phi \lor \varphi \mid X\phi \mid G\phi \mid F\phi \mid \phi \hspace{0.5mm} U \varphi
\end{equation}
Here, $\alpha$ is an atomic proposition mapping an environment state to a Boolean value. Standard logical operators include $\neg$ (negation), $\land$ (conjunction), $\lor$ (disjunction), and $\rightarrow$ (implication). Temporal operators are $X$ (next), $G$ (globally), $F$ (eventually), and $U$ (until).

\smallskip
\textbf{Büchi Automaton:}
Every LTL formula can be represented as a Büchi automaton $\mathcal{B} = (Q, q_0, \Sigma, \delta, \mathcal{F})$, where $Q$ is a finite set of states, $q_0 \in Q$ is the initial state, $\Sigma$ is the input alphabet, $\delta: Q \times 2^\Sigma \rightarrow 2^Q$ is the transition function, and $\mathcal{F} \subseteq Q$ is the set of accepting states.

\begin{figure*}[htp!]
  \centering
  \includegraphics[width=0.75\linewidth]{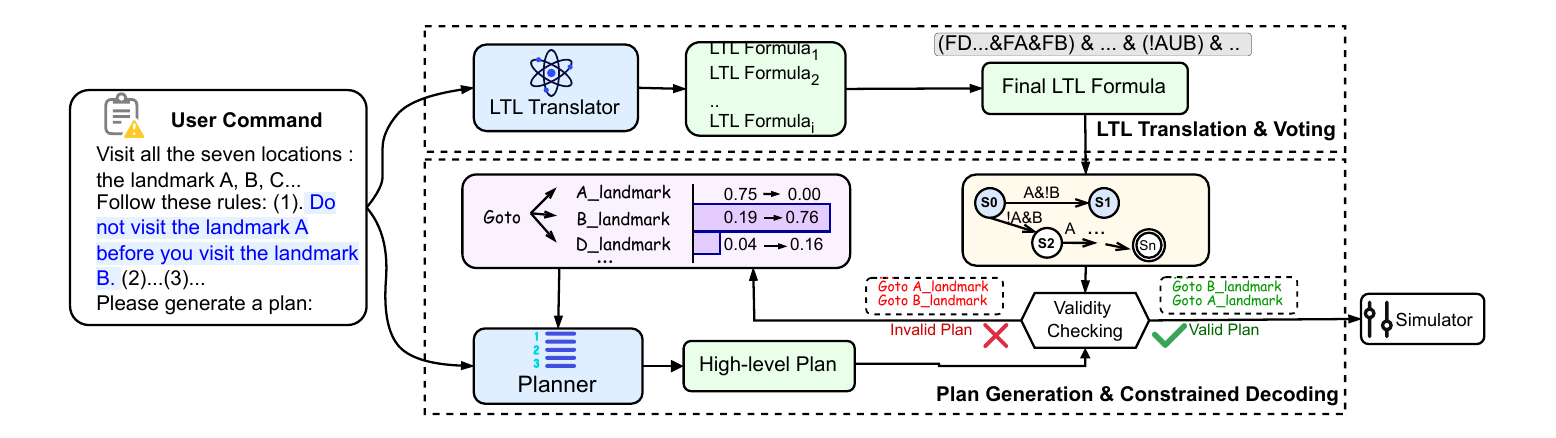}
  \caption{The High-Level Framework of SELP. NL instructions will be input to (a) an LTL translator to build LTL formulas and (b) a planner to generate the probability distribution of each plan step.
    \ourname \emph{enforces} consistency of the generated LTL formulas and the plans sampled from the probability distribution by turning the LTL formula into a \buchi automaton, which monitors and masks out invalid plans. 
    Finally, the plan consistent with the LTL formula will be executed in the simulator.
    }
  \label{fig:framework}
\end{figure*}

\smallskip
\textbf{Task Planning with Co-Safe LTL Constraints:}
Let $S$ be the set of robot states and $A$ be the set of robot actions. Task planning aims to find a sequence of actions $a_1, a_2, \ldots$, where $a_i \in A$, which satisfies the LTL specification. This sequence should generate:
(1) A sequence of robot states $s_0, s_1, s_2, \ldots$ where $s_i \in S$, and
(2) a run of the Büchi automaton $q_0, q_1, q_2, \ldots$ where $q_i \in Q$. The plan must satisfy three conditions:
(1) $
\forall i \geq 0:  s_{i+1} = T(s_i, a_i),
$
(2) $
    \forall i \geq 0: q_{i+1} \in \delta(q_i, L(s_i), 
$
and (3) $
    \exists i \geq 0, \exists q_f \in \mathcal{F}: q_i = q_f.
$
Here, $T: S \times A \rightarrow S$ is the robot's transition function and $L: S \rightarrow 2^\Sigma$ maps robot states to sets of atomic propositions. In practice, we consider co-safe LTL \cite{kupferman2001model} for finite-step planning, and the three conditions are defined in finite runs. This formalization ensures that the robot's behavior, represented by its state sequence, corresponds to an accepting run of an automaton, thus satisfying the co-safe LTL specification. 

\section{APPROACH}

Fig.~\ref{fig:framework} shows an overview of our framework. In this section,
we explain how we collect data (Sec.~\ref{sec:dataset}) for fine-tuning an LLM translator (Sec.~\ref{sec:LTL Translation}) and an LLM planner (Sec.~\ref{sec:plan_finetuning}) to generate time-efficient task plans. %
We design an effective equivalence voting mechanism that enhances NL-LTL translation (Sec.~\ref{sec:LTL Translation}) and a novel LTL-enforced constrained decoding algorithm (Sec.~\ref{sec:LTL Constrained Decoding}) to ensure the safety of generated plans. %

\subsection{Data Collection}
\label{sec:dataset}

\textbf{LTL translation}
To generate diverse NL-LTL pairs for both training and test, we follow~\cite{corl20_ltl} to use context-free grammars to automatically generate LTL formulas. Then, for each LTL formula, we parse its syntax tree and translate it to a structured English description. Since such structured English is monotonous and less natural, we apply GPT-3.5 to paraphrase each generated English description following~\cite{Data-Efficient-Learning}. To create test data, we use GPT-4 to paraphrase the data.

\textbf{Plan Generation}
Given an NL task description $l_{task}$ and an environment description $l_{env}$, the planner model is expected to generate a safe and efficient plan $\mathcal{P}$.
To create training data for LLMs to learn to generate such a plan, we search through safe plans from the automatically produced LTL formulas and then select the most efficient plan based on their simulation time.
Specifically, we combine a navigation task specified by an LTL formula $\phi_0$ with $N$ constraint LTL formulas $\phi_1, ..., \phi_N$, and convert the combined LTL specification $\phi = \land_{i=0}^{N} \phi_i$ into a \buchi automaton $\buchiauto$. Using brute-force search over $\buchiauto$, we generate a set of action sequences $\{P_j\}_{j=1}^{M}$ that result in accepting runs. We then simulate these plans, evaluate their time costs, and choose the most efficient plan for training the LLM Planner.

\subsection{LTL Translation}
\label{sec:LTL Translation}
We finetune an LLM to generate an LTL specification given an NL description. Following prior work~\cite{lang2ltl}, we perform lifted LTL translation for generalization to different environments. 
For example, the NL description ``Head to Walmart and then CVS" will be lifted as ``Head to A and then B", then translated to LTL formula $F(${\small $A$}$\,\&\,F${\small $B$}$)$, and grounded back to $F(${\footnotesize $Walmart$}$\;\&\;F${\footnotesize $CVS$}$)$ with a mapping {\small $\{A\rightarrow Walmart,\,B\rightarrow CVS\}$}. We also adopt LTL formulas in prefix format to avoid parenthesis matching~\cite{lang2ltl}.

\begin{figure*}[htp]
  \centering
  \includegraphics[width=0.75\linewidth]{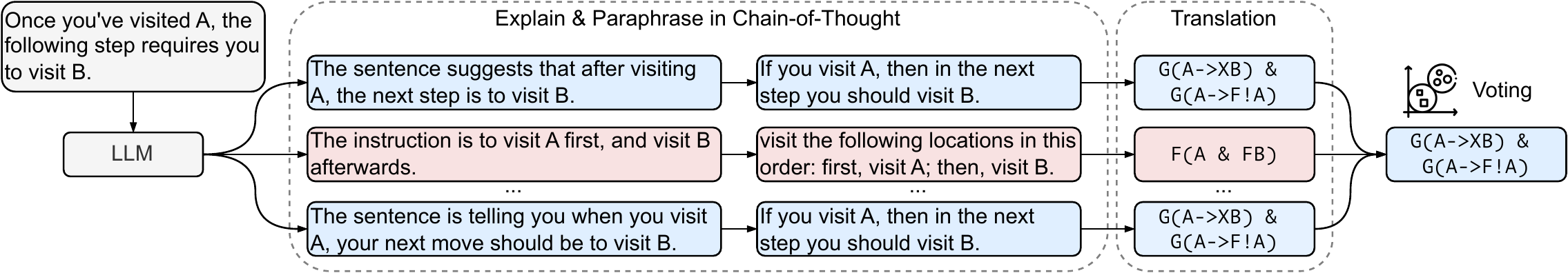}
  \caption{LTL Translation: Equivalence Voting}
  \label{fig:voting}
\end{figure*}

To increase the accuracy of LTL translation, we apply a voting mechanism~\cite{selfconsistency_voting} combined with chain-of-thoughts during LLMs' inference process to accurately capture the temporal logic contained in the user's description. For example, the command in Fig.\ref{fig:voting} implies that if the agent visits A, then in the next step it should visit B. To let the LLM correctly translate such temporal logic, we first prompt the LLM to explain the sentence and paraphrase it in a chain-of-thoughts manner to express the temporal logic explicitly.
Thus, the paraphrased sentence will clearly convey the temporal logic, making it easier for the LLM to comprehend. The fine-tuned LLM takes the paraphrased sentence as input and generates LTL specifications. As NL is diverse, we let the LLM generate 20 explanations and paraphrases  given a user command, for each of which we sample 10 LTL formulas using the fine-tuned LTL model.
These LTL specifications are grouped by their logical equivalency.
We use the \code{spot.are\_equivalent} function in Spot~\cite{duret2016spot} to check if two LTL specifications are equivalent. Finally, the LTL specification from the set with the maximum cardinality is selected as the output.

\subsection{Constrained Decoding for Plan Generation}
\label{sec:LTL Constrained Decoding}
To ensure safety, it is critical that plans should adhere to user-specified constraints.
We propose a constrained decoding algorithm enforced by LTL to prune unsafe actions.
Given an LTL specification $\phi$, its automaton $\buchiauto$, and an input text $I=(l_{env}, l_{task}, a_{1:i-1})$  to an LLM, assume the agent is currently at automaton state $q_{i-1}$ and the LLM generates an action string $a_i$. We check the validity of $a_i$ by progressing over $\buchiauto$ 
to obtain the next automaton state $q_i = \delta(q_{i-1}, a_i) $. If $q_i$ is an invalid automaton state (i.e., there is no path from $q_i$ that leads to any accepting state of $\buchiauto$), then $a_i$ is proved unsafe and we modify LLM's probability distribution by setting the probability of generating $a_i$ to zero, i.e., $p(a_{i}|I)= 0$. In this way, it ensures that the LLM will not generate $a_i$ again for the current step and will ultimately output a safe action by masking out all unsafe actions.

\begin{figure*}[htp]
  \centering
  \includegraphics[width=0.8\linewidth]{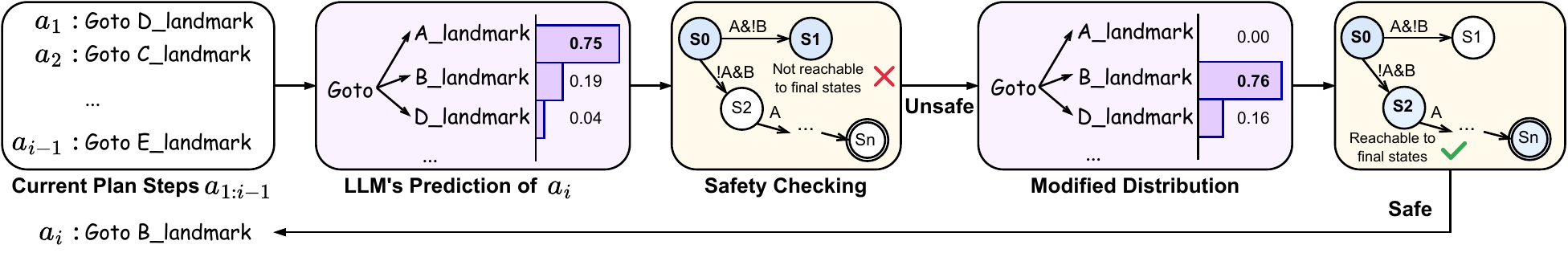}
  \caption{ Example of how an LTL automaton enforces the inference of the LLM planner. An LLM planner (the purple box) predicts the probability of the next tokens, and a Buchi automaton (the yellow box) checks whether these tokens will result in any invalid states and prevent the planner from sampling the tokens that violate constraints.}
  \label{fig:constrained_decoding}
\end{figure*}

Figure~\ref{fig:constrained_decoding} demonstrates an example of the constrained decoding. After generating the first word ``Goto", the LLM will generate the probability distribution for the next token, which represents the location the agent should visit next.  %
The LLM initially generates probabilities of 0.75, 0.19, and 0.04 for ``A\_landmark", ``B\_landmark", and ``D\_landmark" respectively, and samples ``A\_landmark" as the output. The action string ``Goto A\_landmark" is validated through the automaton and turns out to lead to an invalid state $S_1$. Thus, the probability of generating ``A\_landmark" is set to zero and the probability distribution is re-normalized. The LLM resamples on the new probability distribution and generates a new action string: ``Goto B\_landmark", which is proved safe. This iterative process will continue until a valid action string is generated for the current plan step.

\subsection{Fine-Tuning Planner}
\label{sec:plan_finetuning}
Besides safety, the efficiency of task plans is a crucial concern, as it imposes practical constraints on the plan’s viability in real-world scenarios.
This challenge lies outside the capabilities of LTL constraint checking, which solely enforces safety, leaving the demand of efficiency unaddressed.

We introduce a fine-tuning phase for our LLM Planner to bridge this gap, where we finetune an LLM with an efficient plan $P_{e}$ for each task (i.e., the most efficient plan we sampled in Sec.\ref{sec:dataset}). The fine-tuning enables LLMs to learn how to prioritize generating safe plans with the shortest completion time. Assume the textual form of $P_{e}$ is a sequence of $m$ tokens $P_e = \{y_1, y_2, ...., y_m\}$. The training minimizes the negative log-likelihood loss: $L_{LM} = -  \log p(P_{e} |l_{task}, l_{env}) = - \sum_{i=1}^{m} \log p(y_{i} | y_{<i}, l_{task}, l_{env})$.

\section{EXPERIMENTS AND RESULTS}

Our approach, \ourname, introduces three key insights: equivalence voting for robust LTL translation, constrained decoding for safe plan generation, and domain-specific fine-tuning for efficient planning. SELP addresses limitations in handling complex, long-horizon tasks with multiple constraints. Our experiments answer the following research questions:

\noindent\textbf{RQ1:} What is the complexity of our newly introduced datasets compared to existing benchmarks?

\noindent\textbf{RQ2:} What are the safety, completion rate, and execution time of SELP compared to existing LLM planners?

\noindent\textbf{RQ3:} How does equivalence voting improve the accuracy and robustness of LTL translation from natural language?

\noindent\textbf{RQ4:} What impact does constrained decoding have on the safety and efficiency of generated plans?

\noindent\textbf{RQ5:} How does domain-specific fine-tuning enhance the planner's ability to generate efficient plans?

\subsection{Experimental Setup}
\label{sec:Experimental Setup}

\subsubsection{Dataset and Simulation Environment %
}

We create two new datasets for training and evaluation: a drone navigation dataset---\emph{DroneNav}, and a tabletop manipulation dataset---\emph{TabletopManip}. These datasets address the limitations of existing datasets, which are either (1) relatively simple, which are unsuitable for evaluating long-horizon task planning~\cite{code_as_policy, berg2020grounding, nl2ltl_RNN1, Data-Efficient-Learning}, or (2) for common household tasks~\cite{safe-chip, ProgPrompt}, which primarily harness LLMs' common sense for household routines, which is not the focus of this work.

\begin{figure}[htp]
  \centering
  \begin{minipage}[t]{0.24\textwidth}
    \centering
    \includegraphics[width=\linewidth]{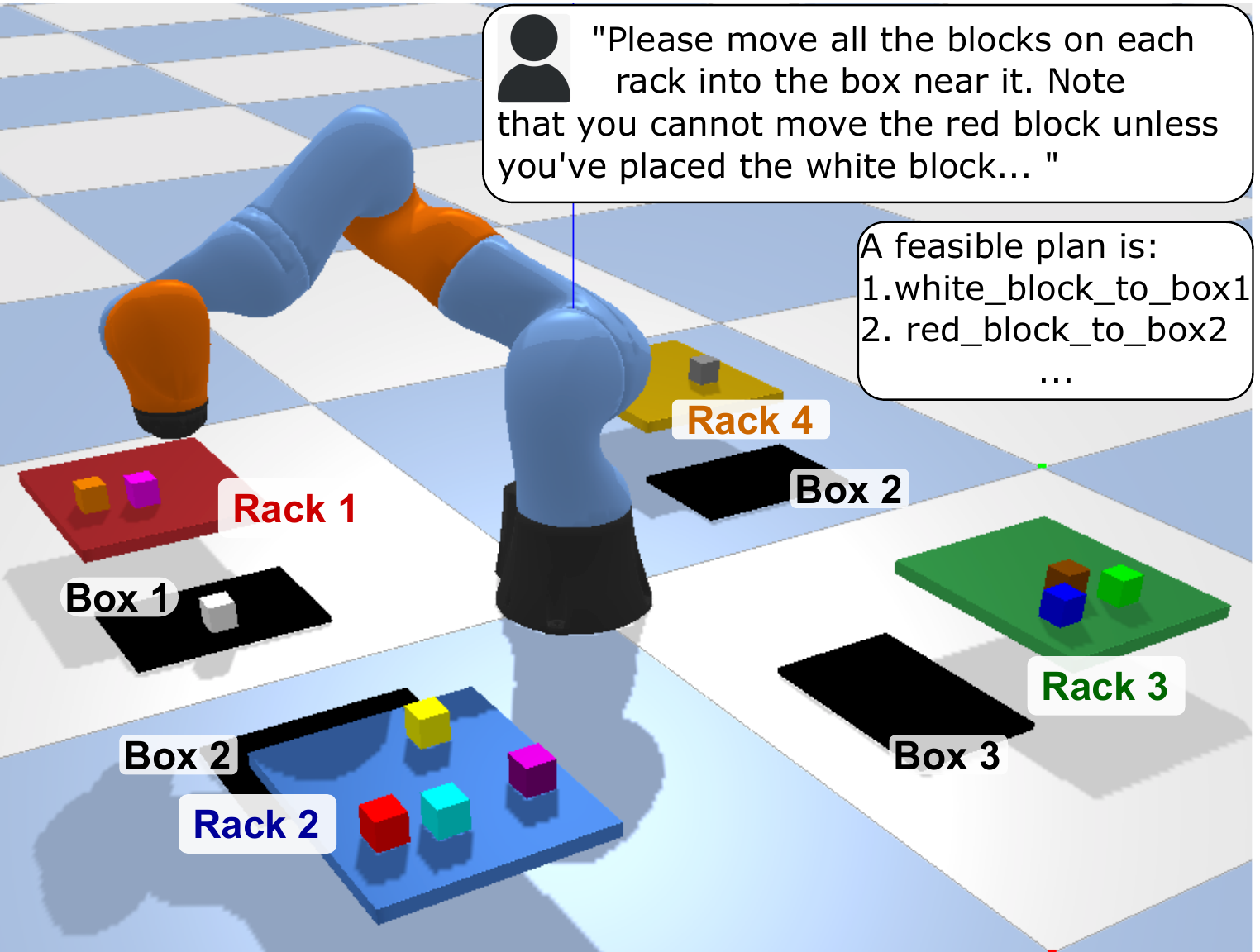}
    \caption{Tabletop Manipulation}
    \label{fig:tabletop_manip_simulation}
  \end{minipage}%
  \hfill
  \begin{minipage}[t]{0.24\textwidth}
    \centering
    \includegraphics[width=\linewidth]{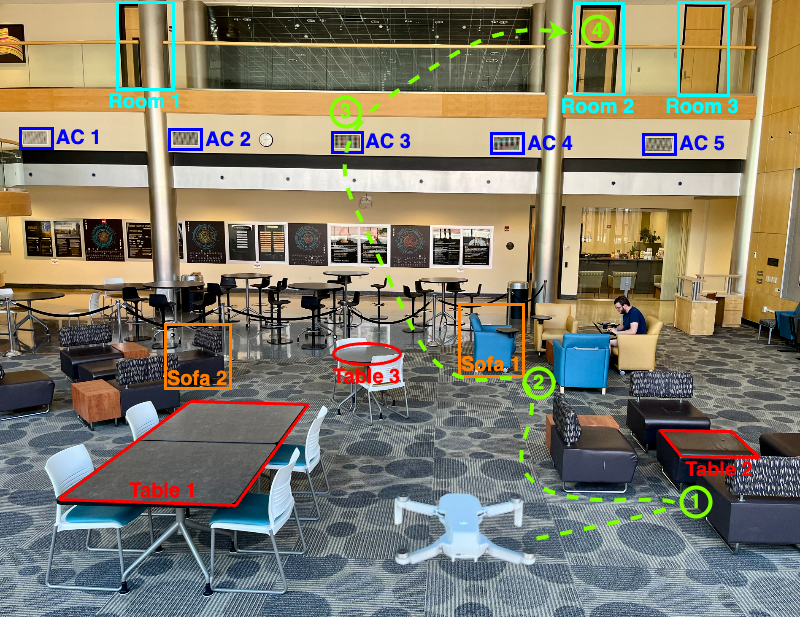}
    \caption{Drone Navigation}
    \label{fig:real-drone-demo}
  \end{minipage}
\end{figure}

\smallskip
\textbf{Dataset complexity (RQ1)}
The syntax trees of LTL specifications~\cite{Data-Efficient-Learning} in DroneNav and TabletopManip have an average depth of 6.89 and 6.71, and an average width of 11.83 and 11.26, respectively, compared to an average depth of 3.46-3.77 and width of 1.78-1.98 in other datasets~\cite{berg2020grounding, nl2ltl_RNN1, Data-Efficient-Learning }. 
To further illustrate the complexity of our dataset, we analyzed the automatons translated from LTL specifications in our dataset. For DroneNav, the average number of nodes and edges of automatons are 354.0 and 21191.5; for TabletopManip, the number of nodes and edges are 338.2 and 26321.6. These two datasets will be released in \href{https://github.com/lt-asset/selp}{github.com/lt-asset/selp}.

i). \underline{\textbf{DroneNav}:} DroneNav consists of navigation tasks requiring an agent to visit a set of locations in a non-predefined order (e.g., visiting all rooms in the building), while conforming to constraints (e.g., the green room must be visited before visiting the yellow room). Each task has multiple feasible plans.
To evaluate the effectiveness of \ourname in planning navigation tasks of varying complexities, we create test data with different numbers (from 1 to 5) of constraints. The drone simulation environment has a three- or four-story building with twelve rooms, as shown on the right side of Fig.~\ref{fig:Motivation_example}. We randomize the room locations and the initial position of the drone to create different environment layouts. We customize this domain in PyBullet Drones~\cite{panerati2021learning} and control the drone to follow the output plan from LLMs with a PID controller. We further deployed the simulation results to the real world, as shown in Fig.~\ref{fig:real-drone-demo} and our video.

ii). \underline{\textbf{TabletopManip}:} TabletopManip dataset instructs an agent to perform pick-and-place tasks, i.e., moving blocks on the racks into designated boxes while conforming to constraints regarding temporal order as shown in Fig.~\ref{fig:tabletop_manip_simulation}. Similar to DroneNav, each task comes with 1 to 5 constraints and has multiple feasible plans. The TabletopManip simulation environment 
has 16 colored blocks on  four racks. The blocks' position and the robot arm's initial position are randomized. We simulate the KUKA LBR iiwa, a 7-DOF collaborative robot arm, using PyBullet. The arm's end-effector is positioned using PyBullet's built-in inverse kinematics solver and controlled by a PID controller.

\subsubsection{\ourname Fine-Tuning}
To build \ourname's LTL translator, we fine-tune CodeLlama2-7b~\cite{codellama2} with 22,662 pairs of NL commands and LTL specifications (Sec.\ref{sec:dataset}). To create \ourname's planner, we fine-tune Llama2-7b~\cite{llama2} with 37,079 pairs of NL commands and plans (Sec.~\ref{sec:dataset}) over 1,159 environments for drone navigation, and 29,037 pairs over 1,000 environments for tabletop manipulation tasks. The test dataset consists of 500 data points across 100 different environment layouts (5 per environment) for both domains. The testing environment layouts are unseen during  training.

\subsubsection{Baseline and SOTA LLM Planners}
We compare  \ourname with the following LLM planners, which all receive the environment description and the NL task description as input:

\smallskip
\textbf{Baseline LLMs} are general-purpose LLMs that are not specialized for robotics tasks. %
We evaluate 3 LLMs---Llama2-7b, GPT-3.5, and GPT-4. We provide three examples of optimal plans to baseline LLMs for in-context learning. We use outputs from these LLMs as task plans.

\smallskip

\textbf{Safety Chip}~\cite{safe-chip} represents NL descriptions into LTL, uses an LLM (GPT-4) to generate plan steps, and verifies each plan step with LTL automatons. Safety Chip is a prompting-based approach that queries the LLM to analyze the violation and regenerates a new plan step iteratively.

 \textbf{Code-as-Policies}~\cite{code_as_policy} is instructed through NL descriptions to generate robot policy code in Python by integrating classic logic structures and 
 third-party libraries (e.g., NumPy).

\subsection{Plan Generation Result (\textbf{RQ2})}
\label{sec:overall_eval}
\label{sec: PlanGenerationResult}

We evaluate \ourname and existing LLM planners with four metrics: safety rate (\textbf{SF}, the percentages of plans that satisfy the task specification), completion rate (\textbf{CP}, the percentages of plans that complete the navigation or manipulation regardless of constraints), plan execution time cost (\textbf{ET}, the average execution time of safe plans in time-steps), and planning time cost (\textbf{PT}, the average planning time in seconds).

We compare \ourname with a brute-force search approach (Sec.~\ref{sec:dataset}) to generate plans from LTL formulas (i.e., LTL+BFS). Since LTL+BFS searches the entire space that conforms to LTL specifications, any plans generated by the brute-force approach are expected to conform to the LTL specifications. However, it may generate inefficient plans. 
We set a time limit of 300 seconds for all techniques except for GPT-4, due to GPT-4's hight cost (instead, the prompting iteration limit for GPT-4 is 30).  %
Timeout is regarded as failures to generate a safe or complete plan.

\begin{table}[t]
    \centering
    \scriptsize
    \setlength{\tabcolsep}{5pt}
    \caption{Drone Navigation and Tabletop Manipulation Experiments. $\uparrow$: higher is better, $\downarrow$: lower is better.}
    \begin{tabular}{@{}l|rrrr|rrrr@{}}
        \toprule
        \multirow{2}{*}{\textbf{Method}} & \multicolumn{4}{c|}{\textbf{Drone Navigation}} & \multicolumn{4}{c}{\textbf{Table-Top Manipulation}}                                                                                                                                                         \\
        \cmidrule(l){2-5} \cmidrule(l){6-9}
                                         & \textbf{SF}$\uparrow$                          & \textbf{CP}$\uparrow$                               & \textbf{ET}$\downarrow$ & \textbf{PT}$\downarrow$ & \textbf{SF}$\uparrow$ & \textbf{CP}$\uparrow$ & \textbf{ET}$\downarrow$ & \textbf{PT}$\downarrow$ \\
        \midrule
        Llama2-7b                        & 5.2                                            & 41.8                                                & 787.9                   & 1.2                     & 3.0                   & 19.0                  & 870.7                   & 3.3                     \\
        GPT-3.5                          & 22.8                                           & 71.4                                                & 739.5                   & 1.1                     & 16.4                  & 46.8                  & 869.1                   & 1.8                     \\
        GPT-4                            & 37.6                                           & 78.2                                                & 725.4                   & 2.2                     & 30.8                  & 66.0                 & 877.5                   & 4.1                     \\
        Code as Policies                 & 10.6                                           & 79.6                                                & 790.0                   & 1.0                     & 10.2                  & 45.4                  & 877.1                   & 2.0                     \\
        Safety Chip                      & 84.4                                           & 91.8                                                & 771.5                   & 28.1                    & 73.2                  & 86.2                  & 819.7                   & 52.3                    \\
        LTL+BFS                          & 84.6                                           & 84.6                                                & 762.3                   & 140.7                   & 74.0                  & 74.0                  & 930.0                   & 138.5                   \\ \midrule
        \ourname-cross                     & 86.6                                             & 93.4                                                  & 701.4                      & 4.8                      & 87.2                  & 90.8                  & 806.7                   & 7.5                     \\
        \textbf{\ourname}                & \textbf{95.2}                                  & \textbf{100.0}                                      & \textbf{581.9}          & 6.4                     & \textbf{93.6}         & \textbf{99.4}         & \textbf{805.2}          & 7.6                     \\
        \bottomrule
    \end{tabular}
    \label{tab:baseline_compare}
\end{table}

Table~\ref{tab:baseline_compare} shows that
\emph{SELP significantly outperforms other techniques in both drone navigation and tabletop manipulation tasks}. For drone navigation, SELP achieves the highest safety rate of 95.2\%, the highest completion rate of 100\%, and the lowest average execution time of 581.89 time-steps. For tabletop manipulation, it achieves the highest safety rate of 93.6\%, the highest completion rate of 99.4\%, and an average execution time of 805.2 time-steps. 
In contrast, baseline LLMs and Code as Policies make poor use of constraints for plan generation, which is reflected in their low safety rates (3.0\%--37.6\%). Safety Chip also uses LTL specifications to enhance safety, achieving a higher safety rate of 84.4\% and 73.2\% for the respective tasks, but still worse than \ourname's performance.
LTL+BFS has a safety rate lower (by 10.6\% and 19.6\%) than \ourname due to timeout. Its ET is 31.0\% and 15.5\% longer than \ourname since it has no knowledge of the environment.
 Compared to the two techniques that are designed to use LTL for plan generation, %
\ourname's PT is a fraction of  Safety Chip's and LTL+BFS's PT, demonstrating it is efficiency in utilizing constraints.
Overall, our results show that by combining equivalence voting, constrained decoding, and fine-tuning, \ourname\ is effective in generating safe and efficient plans.%

In addition, we perform a cross-domain evaluation to assess the generalizability of SELP. Specifically, we fine-tune the LLM planner on drone navigation tasks, while test with tabletop manipulation tasks, and vice versa.%
This scenario applies when resources or training data are limited for finetuning. As shown in Table~\ref{tab:baseline_compare}, SELP-cross can achieve a safety rate of 87.2\%, a completion rate of 90.8\% for tabletop manipulation tasks, a safety rate of 86.6\%, completion rate of 93.4\%, for drone navigation. 
These results suggest that the reasoning ability to solve temporal constraints learned during training by LLMs is transferable. The high safety rates also benefit from SELP's constrained decoding algorithm, which is generalizable to different domains.

\begin{figure}[htp] 
  \centering
  \includegraphics[width=0.85\linewidth]{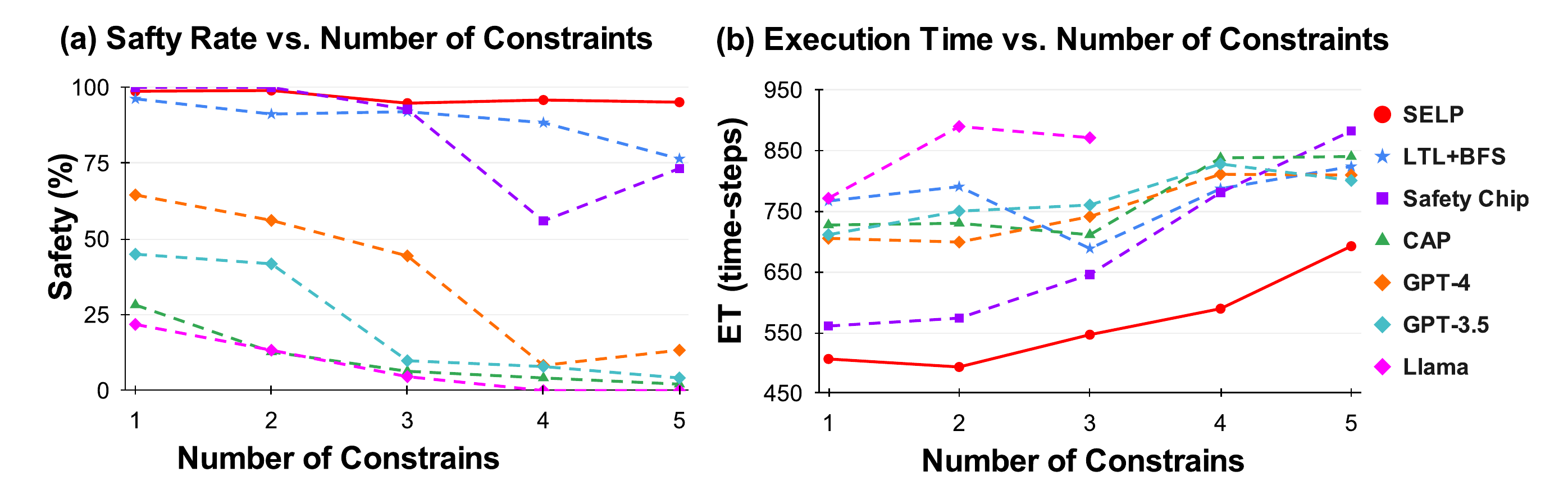}
 \caption{(a) Safety Rate by the Number of Constraints and (b) Execution Time Cost by the Number of Constraints on DroneNav test dataset. No results are shown for Llama for 4 and 5 constraints because Llama fails to generate any safe plans for those tasks.}
\label{fig:analysis_plot}
\end{figure}

We further analyze the plan safety rate and execution time as task complexity increases. %
Fig.~\ref{fig:analysis_plot} (a) shows that \emph{\ourname's improvement over other SOTA LLM planners increases as the tasks become more complex}, demonstrating its ability to generate safe plans for complex tasks. %
Fig.~\ref{fig:analysis_plot} (b) shows \ourname achieves the least execution time across all levels of complexity.

\subsection{LTL Translation Results (\textbf{RQ3})}
\label{sec:ltl_trans_eval}

\begin{table}[h]
    \centering
    \scriptsize
    \caption{LTL Translation Accuracy.}
    \begin{tabular}{l|cccc}
    \hline
    \multicolumn{1}{c}{}    & OSM~\cite{berg2020grounding} & CleanUp~\cite{nl2ltl_RNN1} & DroneNav   & TableManip\\ \hline
    CopyNet~\cite{gu2016incorporating}  & 88.9\%  & 10.4\%    & --   & --  \\
    Lang2LTL~\cite{lang2ltl}    & 94.0\%  & 88.0\%    & 10.8\%  & -- \\
    \begin{tabular}[c]{@{}l@{}}\ourname w/o voting\end{tabular} & 92.7\%  & 91.0\%    & 88.4\% & 87.4\% \\
    \ourname w/ voting  & \textbf{96.6\%} & \textbf{96.0\%}   & \textbf{98.0\%} & \textbf{95.2\%} \\ \hline
    \end{tabular}
    \label{tab:ltl_trans}
\end{table}

We compare \ourname's LTL translation component with two LTL translation approaches, Lang2LTL~\cite{lang2ltl} and CopyNet~\cite{gu2016incorporating}, for navigation tasks. We re-trained Lang2LTL with CodeLlama2-7b for a fair comparison. 
Table~\ref{tab:ltl_trans} shows that SELP's translation module achieves 88.4\% and 98.0\% accuracy without and with equivalence voting on the DroneNav dataset, showing an improvement of 9.8\%. For the TabletopManip dataset, accuracy without and with equivalence voting is 87.4\% and 95.2\%, with an improvement of 7.8\%. We further evaluate our translation module on two crowd-sourced LTL translation datasets: the OSM dataset~\cite{berg2020grounding} and the Cleanup World dataset~\cite{nl2ltl_RNN1}. The result (the first two columns of Table~\ref{tab:ltl_trans}) shows that the voting mechanism consistently improves the accuracy -- by 2.6\% on OSM and by 8.0\% on Cleanup World compared with Lang2LTL, achieving the highest accuracy on both benchmarks.

\subsection{Constrained Decoding and Fine-Tuning (\textbf{RQ4} \& \textbf{RQ5})}

\label{sec:ablation}

\begin{table}[h]

    \scriptsize
    \centering
    \caption{Ablation Study. \textit{FT} means finetuning with plan data; \textit{CD} means constrained decoding. 
    } %
    \label{tab:abotione}
    \addtolength{\tabcolsep}{-0.3em}
    \begin{tabular}{l|rrrr|rrrr}
    \toprule Task    & \multicolumn{4}{c|}{Drone Navigation}   & \multicolumn{4}{c}{Tabletop Manipulation}   \\ \midrule
    Metrics      & \textbf{SF} $\uparrow$   & \textbf{CP} $\uparrow$ & \textbf{ET} $\downarrow$ & \textbf{PT} $\downarrow$ & \textbf{SF} $\uparrow$   & \textbf{CP} $\uparrow$ & \textbf{ET} $\downarrow$ & \textbf{PT} $\downarrow$ \\ \midrule
    \textbf{\ourname} w/o \textit{FT}  & 77.8  & 84.6 & 968.0 & 10.1  & 81.4 & 87.0   & 878.5  & 11.9   \\
    \textbf{\ourname}  w/o \textit{CD} & 80.8   & 95.4  & \textbf{541.0}  & 2.9  & 69.6 & 88.4   & 816.9  & 3.5   \\
    \textbf{\ourname}    & \textbf{95.2}  & \textbf{100.0}   & 581.9 & 6.4  &\textbf{93.6}  & \textbf{99.4} & \textbf{805.2}   & 7.6  \\
    \bottomrule
    \end{tabular} \\
\end{table}

Table~\ref{tab:abotione} presents the ablation study results, comparing \ourname with and without finetuning with plan data (FT) and constrained decoding (CD). \ourname w/o FT shows lower safety, and completeness, with a significantly longer execution time, which justifies the necessity of fine-tuning. \ourname outperforms \ourname w/o CD in safety, and completeness. However, for drone navigation, \ourname has a slightly longer ET than \ourname w/o CD due to that CD reshapes the probability distribution of the LLM planner to ensure safety. This degeneration in ET (-7.6\%) is acceptable considering the importance of achieving a higher safety rate (+14.4\%). \ourname, with both FT and CD, achieves the highest performance in terms of safety and completeness.

\section{CONCLUSION AND FUTURE WORK}

We propose a novel approach \ourname to generate both safe and efficient plans. \ourname consists of an LTL translator with a voting mechanism and a fine-tuned planner with a constrained decoding algorithm.
Our experiment on drone navigation and tabletop manipulation tasks demonstrate:
(1) the voting mechanism effectively improves LTL translation accuracy 
(2) the constrained decoding algorithm is critical for ensuring safety
(3) domain-specific fine-tuning is essential to adapt LLM to generate efficient plans as well as meet safety standards. \ourname outperforms SOTA LLM planners and is generablizable to different domains. 

We note a few limitations: (1) we only consider generating task plans with finite steps despite LTL's capability to express plans with infinite steps; (2) we do not design a feedback mechanism. However, it is straightforward to re-generate a plan when SELP fails to produce a safe plan or when the generated plan is inconsistent with the NL task description. For future work, we will focus on the following aspects: (1) consider more objectives such as energy consumption. (2) include vision information and expand on multi-modality.

\section*{Acknowledgement}
This research was supported in part by NSF 1901242 and 2006688. Any opinions, findings, and conclusions in this paper are those of the authors only and do not necessarily reflect the views of our sponsors.

\bibliographystyle{IEEEtran}
\bibliography{root}

\begin{thebibliography}{10}
\providecommand{\url}[1]{#1}
\csname url@rmstyle\endcsname
\providecommand{\newblock}{\relax}
\providecommand{\bibinfo}[2]{#2}
\providecommand\BIBentrySTDinterwordspacing{\spaceskip=0pt\relax}
\providecommand\BIBentryALTinterwordstretchfactor{4}
\providecommand\BIBentryALTinterwordspacing{\spaceskip=\fontdimen2\font plus
\BIBentryALTinterwordstretchfactor\fontdimen3\font minus \fontdimen4\font\relax}
\providecommand\BIBforeignlanguage[2]{{%
\expandafter\ifx\csname l@#1\endcsname\relax
\typeout{** WARNING: IEEEtran.bst: No hyphenation pattern has been}%
\typeout{** loaded for the language `#1'. Using the pattern for}%
\typeout{** the default language instead.}%
\else
\language=\csname l@#1\endcsname
\fi
#2}}

\bibitem{code_as_policy}
J.~Liang, W.~Huang, F.~Xia, P.~Xu, K.~Hausman, B.~Ichter, P.~Florence, and A.~Zeng, ``Code as policies: Language model programs for embodied control,'' in \emph{2023 IEEE International Conference on Robotics and Automation (ICRA)}, 2023, pp. 9493--9500.

\bibitem{saycan}
M.~Ahn, A.~Brohan, N.~Brown, Y.~Chebotar, O.~Cortes, B.~David, C.~Finn, C.~Fu, K.~Gopalakrishnan, K.~Hausman, \emph{et~al.}, ``Do as i can, not as i say: Grounding language in robotic affordances,'' \emph{arXiv preprint arXiv:2204.01691}, 2022.

\bibitem{ProgPrompt}
I.~Singh, V.~Blukis, A.~Mousavian, A.~Goyal, D.~Xu, J.~Tremblay, D.~Fox, J.~Thomason, and A.~Garg, ``Progprompt: Generating situated robot task plans using large language models,'' in \emph{2023 IEEE International Conference on Robotics and Automation (ICRA)}, 2023, pp. 11\,523--11\,530.

\bibitem{yang2023survey}
Z.~Yang, X.~Jia, H.~Li, and J.~Yan, ``Llm4drive: A survey of large language models for autonomous driving,'' 2023.

\bibitem{jiang2022vima}
Y.~Jiang, A.~Gupta, Z.~Zhang, G.~Wang, Y.~Dou, Y.~Chen, L.~Fei-Fei, A.~Anandkumar, Y.~Zhu, and L.~Fan, ``Vima: General robot manipulation with multimodal prompts,'' in \emph{NeurIPS 2022 Foundation Models for Decision Making Workshop}, 2022.

\bibitem{driess2023palm}
D.~Driess, F.~Xia, M.~S. Sajjadi, C.~Lynch, A.~Chowdhery, B.~Ichter, A.~Wahid, J.~Tompson, Q.~Vuong, T.~Yu, \emph{et~al.}, ``Palm-e: An embodied multimodal language model,'' \emph{arXiv preprint arXiv:2303.03378}, 2023.

\bibitem{liu2023llm}
H.~Liu, Y.~Zhu, K.~Kato, I.~Kondo, T.~Aoyama, and Y.~Hasegawa, ``Llm-based human-robot collaboration framework for manipulation tasks,'' \emph{arXiv preprint arXiv:2308.14972}, 2023.

\bibitem{huang2023reasoning}
J.~Huang and K.~C.-C. Chang, ``Towards reasoning in large language models: A survey,'' 2023.

\bibitem{zhu2024knowagent}
Y.~Zhu, S.~Qiao, Y.~Ou, S.~Deng, N.~Zhang, S.~Lyu, Y.~Shen, L.~Liang, J.~Gu, and H.~Chen, ``Knowagent: Knowledge-augmented planning for llm-based agents,'' 2024.

\bibitem{li2024llms}
Z.~Li, Y.~Cao, X.~Xu, J.~Jiang, X.~Liu, Y.~S. Teo, S.~wei Lin, and Y.~Liu, ``Llms for relational reasoning: How far are we?'' 2024.

\bibitem{safe-chip}
Z.~Yang, S.~S. Raman, A.~Shah, and S.~Tellex, ``Plug in the safety chip: Enforcing constraints for llm-driven robot agents,'' \emph{arXiv preprint arXiv:2309.09919}, 2023.

\bibitem{huang2022language}
W.~Huang, P.~Abbeel, D.~Pathak, and I.~Mordatch, ``Language models as zero-shot planners: Extracting actionable knowledge for embodied agents,'' \emph{arXiv preprint arXiv:2201.07207}, 2022.

\bibitem{vemprala2023chatgpt}
S.~Vemprala, R.~Bonatti, A.~Bucker, and A.~Kapoor, ``Chatgpt for robotics: Design principles and model abilities. 2023,'' 2023.

\bibitem{huang2022inner}
W.~Huang, F.~Xia, T.~Xiao, H.~Chan, J.~Liang, P.~Florence, A.~Zeng, J.~Tompson, I.~Mordatch, Y.~Chebotar, \emph{et~al.}, ``Inner monologue: Embodied reasoning through planning with language models,'' \emph{arXiv preprint arXiv:2207.05608}, 2022.

\bibitem{shah2023lm}
D.~Shah, B.~Osi{\'n}ski, S.~Levine, \emph{et~al.}, ``Lm-nav: Robotic navigation with large pre-trained models of language, vision, and action,'' in \emph{Conference on robot learning}.\hskip 1em plus 0.5em minus 0.4em\relax PMLR, 2023, pp. 492--504.

\bibitem{hazra2024saycanpay}
R.~Hazra, P.~Z. Dos~Martires, and L.~De~Raedt, ``Saycanpay: Heuristic planning with large language models using learnable domain knowledge,'' in \emph{Proceedings of the AAAI Conference on Artificial Intelligence}, vol.~38, no.~18, 2024, pp. 20\,123--20\,133.

\bibitem{tidybot}
J.~Wu, R.~Antonova, A.~Kan, M.~Lepert, A.~Zeng, S.~Song, J.~Bohg, S.~Rusinkiewicz, and T.~Funkhouser, ``Tidybot: Personalized robot assistance with large language models,'' \emph{Autonomous Robots}, vol.~47, no.~8, pp. 1087--1102, 2023.

\bibitem{autotamp}
Y.~Chen, J.~Arkin, C.~Dawson, Y.~Zhang, N.~Roy, and C.~Fan, ``Autotamp: Autoregressive task and motion planning with llms as translators and checkers,'' in \emph{2024 IEEE International Conference on Robotics and Automation (ICRA)}.\hskip 1em plus 0.5em minus 0.4em\relax IEEE, 2024, pp. 6695--6702.

\bibitem{xie2023translating}
Y.~Xie, C.~Yu, T.~Zhu, J.~Bai, Z.~Gong, and H.~Soh, ``Translating natural language to planning goals with large-language models,'' \emph{arXiv preprint arXiv:2302.05128}, 2023.

\bibitem{liu2023llm+}
B.~Liu, Y.~Jiang, X.~Zhang, Q.~Liu, S.~Zhang, J.~Biswas, and P.~Stone, ``Llm+ p: Empowering large language models with optimal planning proficiency,'' \emph{arXiv preprint arXiv:2304.11477}, 2023.

\bibitem{berg2020grounding}
M.~Berg, D.~Bayazit, R.~Mathew, A.~Rotter-Aboyoun, E.~Pavlick, and S.~Tellex, ``Grounding language to landmarks in arbitrary outdoor environments,'' in \emph{2020 IEEE International Conference on Robotics and Automation (ICRA)}.\hskip 1em plus 0.5em minus 0.4em\relax IEEE, 2020, pp. 208--215.

\bibitem{nl2ltl_RNN1}
N.~Gopalan, D.~Arumugam, L.~Wong, and S.~Tellex, ``Sequence-to-sequence language grounding of non-markovian task specifications,'' in \emph{Proceedings of Robotics: Science and Systems}, Pittsburgh, Pennsylvania, June 2018.

\bibitem{oh2019planning}
Y.~Oh, R.~Patel, T.~Nguyen, B.~Huang, E.~Pavlick, and S.~Tellex, ``Planning with state abstractions for non-markovian task specifications,'' \emph{arXiv preprint arXiv:1905.12096}, 2019.

\bibitem{Data-Efficient-Learning}
J.~Pan, G.~Chou, and D.~Berenson, ``Data-efficient learning of natural language to linear temporal logic translators for robot task specification,'' \emph{arXiv preprint arXiv:2303.08006}, 2023.

\bibitem{lang2ltl}
J.~X. Liu, Z.~Yang, I.~Idrees, S.~Liang, B.~Schornstein, S.~Tellex, and A.~Shah, ``Grounding complex natural language commands for temporal tasks in unseen environments,'' in \emph{Conference on Robot Learning}.\hskip 1em plus 0.5em minus 0.4em\relax PMLR, 2023, pp. 1084--1110.

\bibitem{chen2023nl2tl}
Y.~Chen, R.~Gandhi, Y.~Zhang, and C.~Fan, ``Nl2tl: Transforming natural languages to temporal logics using large language models,'' \emph{arXiv preprint arXiv:2305.07766}, 2023.

\bibitem{LTLapp1}
A.~Shah, S.~Li, and J.~Shah, ``Planning with uncertain specifications (puns),'' \emph{IEEE Robotics and Automation Letters}, vol.~5, no.~2, pp. 3414--3421, 2020.

\bibitem{LTLapp2}
J.~{Xinyu Liu}, A.~{Shah}, E.~{Rosen}, G.~{Konidaris}, and S.~{Tellex}, ``{Skill Transfer for Temporally-Extended Task Specifications},'' \emph{arXiv e-prints}, p. arXiv:2206.05096, June 2022.

\bibitem{DeGiacomo2018FoundationsFR}
\BIBentryALTinterwordspacing
G.~D. Giacomo, L.~Iocchi, M.~Favorito, and F.~Patrizi, ``Foundations for restraining bolts: Reinforcement learning with ltlf/ldlf restraining specifications,'' in \emph{International Conference on Automated Planning and Scheduling}, 2018. [Online]. Available: \url{https://api.semanticscholar.org/CorpusID:109929091}
\BIBentrySTDinterwordspacing

\bibitem{kupferman2001model}
O.~Kupferman and M.~Y. Vardi, ``Model checking of safety properties,'' in \emph{International Conference on Computer Aided Verification}.\hskip 1em plus 0.5em minus 0.4em\relax Springer, 2001, pp. 172--183.

\bibitem{corl20_ltl}
\BIBentryALTinterwordspacing
C.~Wang, C.~Ross, B.~Katz, and A.~Barbu, ``Learning a natural-language to ltl executable semantic parser for grounded robotics,'' in \emph{Conference on Robot Learning}, 2020. [Online]. Available: \url{https://api.semanticscholar.org/CorpusID:221083453}
\BIBentrySTDinterwordspacing

\bibitem{selfconsistency_voting}
\BIBentryALTinterwordspacing
X.~Wang, J.~Wei, D.~Schuurmans, Q.~V. Le, E.~H. Chi, S.~Narang, A.~Chowdhery, and D.~Zhou, ``Self-consistency improves chain of thought reasoning in language models,'' in \emph{The Eleventh International Conference on Learning Representations}, 2023. [Online]. Available: \url{https://openreview.net/forum?id=1PL1NIMMrw}
\BIBentrySTDinterwordspacing

\bibitem{duret2016spot}
A.~Duret-Lutz, A.~Lewkowicz, A.~Fauchille, T.~Michaud, E.~Renault, and L.~Xu, ``Spot 2.0—a framework for ltl and-automata manipulation,'' in \emph{International Symposium on Automated Technology for Verification and Analysis}.\hskip 1em plus 0.5em minus 0.4em\relax Springer, 2016, pp. 122--129.

\bibitem{panerati2021learning}
J.~Panerati, H.~Zheng, S.~Zhou, J.~Xu, A.~Prorok, and A.~P. Schoellig, ``Learning to fly---a gym environment with pybullet physics for reinforcement learning of multi-agent quadcopter control,'' in \emph{2021 IEEE/RSJ International Conference on Intelligent Robots and Systems (IROS)}, 2021, pp. 7512--7519.

\bibitem{codellama2}
B.~Roziere, J.~Gehring, F.~Gloeckle, S.~Sootla, I.~Gat, X.~E. Tan, Y.~Adi, J.~Liu, T.~Remez, J.~Rapin, \emph{et~al.}, ``Code llama: Open foundation models for code,'' \emph{arXiv preprint arXiv:2308.12950}, 2023.

\bibitem{llama2}
H.~Touvron, L.~Martin, K.~Stone, P.~Albert, A.~Almahairi, Y.~Babaei, N.~Bashlykov, S.~Batra, P.~Bhargava, S.~Bhosale, \emph{et~al.}, ``Llama 2: Open foundation and fine-tuned chat models,'' \emph{arXiv preprint arXiv:2307.09288}, 2023.

\bibitem{gu2016incorporating}
J.~Gu, Z.~Lu, H.~Li, and V.~O. Li, ``Incorporating copying mechanism in sequence-to-sequence learning,'' \emph{arXiv preprint arXiv:1603.06393}, 2016.

\end{thebibliography}

\end{document}